\documentclass[lettersize,journal]{IEEEtran}
\usepackage{amsmath,amsfonts}
\usepackage{array}
\usepackage[caption=false,font=normalsize,labelfont=sf,textfont=sf]{subfig}
\usepackage{textcomp}
\usepackage{stfloats}
\usepackage{url}
\usepackage{verbatim}
\usepackage{graphicx}
\usepackage{cleveref}
\usepackage{cite}

\usepackage{xcolor}

\usepackage{amsmath}
\usepackage{algorithm}
\usepackage{algorithmic}
\DeclareMathOperator*{\argmax}{arg\,max}

\begin{document}

\title{Sparsity-based Safety Conservatism for Constrained Offline Reinforcement Learning}

\author{Minjae Cho\thanks{Minjae Cho is with Department of Mechanical Engineering, Mississippi State University (e-mail: mc3216@msstate.edu)} and
Chuangchuang Sun \thanks{Chuandchuang Sun is with Department of Aerospace Engineering, Mississippi State University (e-mail: csun@ae.msstate.edu)}}

\markboth{Journal of \LaTeX\ Class Files,~Vol.~14, No.~8, August~2021}%
{Shell \MakeLowercase{\textit{et al.}}: A Sample Article Using IEEEtran.cls for IEEE Journals}

\IEEEpubid{0000--0000/00\$00.00~\copyright~2021 IEEE}

\maketitle

\begin{abstract}
Reinforcement Learning (RL) has made notable success in decision-making fields like healthcare, autonomous driving, and robotic manipulation. Yet, its reliance on real-time feedback from the environment poses challenges in costly or hazardous settings. Furthermore, RL's training approach, centered on "on-policy" sampling assumptions, doesn't fully capitalize on previously gathered data. Hence, Offline RL has emerged as a compelling alternative, particularly in scenarios where conducting additional experiments is impractical and abundant datasets are available for leverage. However, the well-documented challenge of distributional shift (extrapolation), indicating the disparity between data distributions and learning policies, also poses a risk in offline RL, potentially leading to significant safety breaches due to estimation errors (interpolation). This concern is particularly pronounced in safety-critical domains, where real-world problems are prevalent. To address both extrapolation and interpolation errors, numerous studies have introduced additional constraints to confine policy behavior, steering it towards a more cautious decision-making regime. While many studies have delved into addressing extrapolation errors \cite{kumar2020conservative, nachum2019algaedice, yu2020mopo}, fewer have focused on providing effective solutions for tackling interpolation errors. For example, \cite{chow2015risk} and \cite{lee2022coptidice}, tackle this issue by incorporating potential cost-maximizing optimization by perturbing the original dataset. However, this, involving a bi-level optimization structure, may introduce significant instability or complicate problem-solving in high-dimensional tasks, calling for an effective method to induce conservatism. This motivates us to pinpoint areas where hazards may be more prevalent than initially estimated based on the sparsity of available data by providing significant insight into constrained offline RL. In this paper, we present conservative metrics based on data sparsity that demonstrate the high generalizability to any methods and efficacy compared to using bi-level cost-ub-maximization. 

\end{abstract}


\section{Introduction}
Reinforcement Learning (RL) has demonstrated remarkable success in decision-making across diverse domains, including healthcare, autonomous driving, and robotic manipulation. The development of efficient learning algorithms has empowered RL systems to extract valuable insights and learn optimal strategies through interactions with their environments, yielding substantial success. However, the current RL paradigm, reliant on an impractical number of experiments, encounters difficulties in environments where real-time feedback is either costly or hazardous. This limitation impedes the practical deployment of RL in many real-world scenarios.

A significant constraint of RL stems from its dependence on "on-policy" sampling assumptions during training, limiting its efficacy in actively sampling domains, despite the abundance of pre-collected data in most settings. To mitigate this constraint, Offline RL has garnered attention as an alternative approach, especially in situations where conducting further experiments is impractical, yet substantial datasets are accessible for learning. While the transition towards an offline learning paradigm presents promising prospects, it also introduces fresh challenges, prominently the concern of estimation errors.

Estimation errors present a substantial risk in offline RL, as they result in inaccurate predictions for unseen state-action pairs, often leading to overestimation of less favorable areas. This phenomenon, known as distributional shift, highlights disparities between the distributions of data and learning policies. Distributional shift encompasses both extrapolation and interpolation errors, both of which fall under the category of estimation errors. While previous methods have effectively addressed the avoidance of extrapolation While \cite{kumar2020conservative, nachum2019algaedice, yu2020mopo}, interpolation errors persist as a critical concern, especially in safety-critical domains, where the ramifications of such errors can be severe.

Researchers have pursued avenues to address distributional shifts, employing constraint methods that impose additional restrictions to steer policies towards more conservative behavior. While considerable attention has been directed toward mitigating extrapolation errors, fewer studies have focused on effectively combating interpolation errors. Some efforts have been made to address interpolation errors in safety-critical domains by integrating potential cost-maximizing optimization techniques, such as perturbing the original dataset \cite{chow2015risk, lee2022coptidice}. However, these methods frequently entail intricate bi-level optimization structures, which may introduce instability or complicate problem-solving, particularly in high-dimensional tasks. Furthermore, these methods do not demonstrate compliance with the specified safety thresholds, merely increasing cost estimates for potentially hazardous regions based on estimations. As a result, there is a pressing need for more efficient approaches to incorporate conservatism into offline RL systems.

Motivated by these challenges, this paper aims to identify areas where hazards may be more prevalent than initially estimated based on the sparsity of available data. We present extensive empirical findings that demonstrate the effectiveness of our approach compared to traditional bi-level cost upper-bound maximization methods. Through this research, we aim to achieve: 1) an efficient safe-conservatism method, and 2) generalizability that is applicable to any algorithmic setting.

\subsection*{Organization}
This paper is structured as follows: In \Cref{sec:preliminaries}, we provide essential insights into offline RL and the computation of sparsity measures using K-means clustering. For readers familiar with these topics, we direct their attention to the methodology section (\Cref{sec:methodology}), where we elaborate on the development of K-means clustering into a sparsity conservatism measure and its associated benefits. Subsequently, in \Cref{sec:experiments}, we demonstrate the effectiveness of our approach through empirical experiments conducted in random constrained Markov decision processes for discrete environment settings and CartPole using \cite{dulacarnold2020realworldrlempirical} for continuous robotic tasks. Finally, we conclude with a discussion and promising avenues for future research.

\section{Preliminaries} \label{sec:preliminaries}

\subsection{Constrained Offline Reinforcement Learning}
Constrained Markov Decision Process (CMDP) \cite{altman-constrainedMDP} considers additional Cost function $C(s,a)$ into an MDP representation encapsulated by $M = \langle \mathcal{S}, \mathcal{A}, T, R, C, p_0, \gamma \rangle$. Here, $\mathcal{S}$ denotes the state set, $\mathcal{A}$ denotes the action set, $R : \mathcal{S} \times \mathcal{A} \rightarrow \mathbb{R}$ and $C : \mathcal{S} \times \mathcal{A} \rightarrow \mathbb{R}$ are the reward function and cost function respectively with $T : \mathcal{S} \times \mathcal{A} \rightarrow \mathcal{S^{'}}$ represents the transition probabilities. Additionally, $p_0 \subset \mathcal{S}_0$ characterizes the initial state distribution, while $\gamma \subset [0, 1]$ stands as the discount factor. The policy $\pi : \mathcal{S} \rightarrow \mathcal{A}$ plays a central role, mapping states to action distributions. Our objective here is to maximize this mapping, thereby enhancing the cumulative reward while meeting specified safety constraints. 

Constrained online RL, as exemplified by \cite{achiam2017constrained} and \cite{yang2020projection}, rely on ``\emph{on-policy}" samples, updating the policy with samples collected under the currently updating policy. However, Constrained Offline Reinforcement Learning (CORL), as demonstrated in the work by \cite{lee2022coptidice}, entails the agent's learning of an optimal policy using a fixed static dataset (``\emph{off-policy samples}'') $\mathcal{D} = \bigl\{ (s_i,a_i,r_i,c_i,s'_i ) \bigl\}^N_{i=1}$ without any knowledge of how the dataset was initially generated. This transition from an online to offline paradigm introduces several challenges that necessitate: 1) intricate hyperparameter adjustments to mitigate overfitting or underfitting, 2) effective generalization to capture the broader environmental dynamics, and 3) robust techniques for Off-Policy Evaluation (OPE) for policy updates with minimal violations of estimations.

\subsection{Off-Policy Evaluation}
OPE involves estimating the expected performance of a learning policy based on historical experiences stored in dataset $\mathcal{D}$ to maximize estimated cumulative returns. Therefore, OPE fulfills several crucial roles, including furnishing high-confidence guarantees prior to deployment and aiding in hyperparameter tuning to evade suboptimal solutions. The sensitivity to factors like learning rates and the level of conservatism regarding Out-of-Distribution (OOD) or in-distribution underscores the importance of effective OPE methods for policy validation. While multiple OPE approaches are available, our focus will be on Importance Sampling (IS) due to its conciseness and relevance to both our theoretical and experimental framework.

\subsubsection{Importance Sampling}
IS is an OPE method that gauges the effectiveness of a learning policy as a value function by approximating the probability ratio between the behavior policy $\pi_\beta$ and our training policy $\pi$. The value estimation of $\pi$ is articulated as follows:
\begin{equation}
    V_\pi = \mathbb{E}_{(s,a) \sim d^{\pi_\beta}} \biggr[ \omega_{0:H} \sum_{t=0}^H \gamma^tr(s_t,a_t) \biggr]
\end{equation}
Here, $\omega_{i:j} = (\prod_{t=i}^j \pi(a_t|s_t))/(\prod_{t=i}^j\pi_\beta(a_t,s_t))$ represents the product of importance weights and accounts for the ratio of probabilities under policies $\pi$ and $\pi_\beta$ for a given trajectory. Intuitively, this estimation involves computing the cumulative rewards multiplied by the disparity in visitation frequencies for each state-action pair in the dataset. However, this product's inclusion over the entire horizon (0 to $H$) can introduce high variance in the correction term $\omega$, resulting in ratios that are so close to zero that they are practically indistinguishable. To address this, variance reduction techniques like doubly robust and marginalized IS, embraced by the DICE-family (DIstribution Correction Estimation) algorithms, have emerged. These methods have excelled in various domains due to their low variance and solid performance by estimating the stationary distribution $d^\pi$ through marginalized probability ratios, where $d^\pi$ is defined as:
\begin{equation}
  d^\pi(s,a)=
  \begin{cases}
    \lim_{T\to\infty} \frac{1}{T+1} \sum^T_{t=0}\text{Pr}(s_t,a_t), & \text{if $\gamma=1$}.\\
    (1-\gamma) \sum^\infty_{t=0} \gamma^t \text{Pr}(s_t,a_t), & \text{if $\gamma < 1$}.
  \end{cases}
\end{equation}

The DICE algorithms with additional cost consideration aim to maximize the expected return of policy $\pi$ through the following optimization shape:
\begin{equation*}
    \begin{aligned}
        & \max \quad \mathbb{E}_{(s,a) \sim d^D} [R(s,a)\omega(s,a)] \\
        & \quad \textrm{s.t.} \quad \mathbb{E}_{(s,a) \sim d^D} [C_k(s,a)\omega(s,a)] \leq \hat{c}_k\\ 
        & \sum_{a'} d(s',a') = (1-\gamma)p_0(s')+\gamma\sum_{s,a}d^D(s,a)T(s'|s,a)
    \end{aligned}
\end{equation*}
Here, $\omega(s,a) = \frac{d^\pi(s,a)}{d^D(s,a)}$ (marginalized probability ratios), and the term $\mathbb{E}_{(s,a) \sim d^D} [R(s,a)\omega(s,a)]$ is therefore the estimated return of policy $\pi$: $[R(s,a)d^\pi]$. This entire process relies exclusively on the state-action distribution within the provided dataset $D$ and the estimation of $\omega(s,a)$.

\subsubsection{COptiDICE: Constrained Offline Policy Optimization via DICE}
COptiDICE is a constrained offline RL algorithm that directly estimates the stationary distribution corrections of the optimal policy by preventing both OOD and in-distributional actions:
\begin{equation}
    \begin{aligned}
        & \max \quad \mathbb{E}_{(s,a) \sim d^D} [R(s,a) - \alpha D_f(d^\pi || d^D)] \\
        & \quad \textrm{s.t.} \quad \mathbb{E}_{(s,a) \sim d^D} [C_k(s,a)\omega(s,a)\omega^{*}(s_0,s,a,s')] \leq \hat{c}_k\\ 
        & \sum_{a^{'}} d(s^{'},a^{'}) = (1-\gamma)p_0(s^{'})+\gamma\sum_{s,a}d^D(s,a)T(s^{'}|s,a)
    \end{aligned}
    \label{eqn:coptidice}
\end{equation}
where $\alpha D_f(d^\pi || d^D)]$ is restricts OOD actions scaled by $\alpha$ and $\omega^{*}(s_0,s,a,s')$ is normalized cost upper bound estimation to further bring conservatism within in-distributional actions. It has demonstrated competitive performance in Tabular MDP and robotic control tasks in safety-constrained settings when compared to other state-of-the-art methods.

\subsection{K-mean Clusterings}
K-means clustering is a popular unsupervised machine learning technique used in data science. The goal of K-Means is to partition a dataset into \(K\) clusters, where each data point belongs to the cluster with the nearest mean (centroid). The algorithm aims to minimize the variance within each cluster. Mathematically, it can be expressed as follows:
\begin{equation} \label{eqn:clustering}
    \underset{G}{\text{argmin}} \sum_{i=1}^{K} \sum_{\mathbf{x} \in G_i} ||\mathbf{x} - \boldsymbol{\mu}_i||^2
\end{equation}
where \(G\) is the clustering solution, \(\boldsymbol{\mu}_i\) is the centroid of cluster \(i\), and \(||\mathbf{x} - \boldsymbol{\mu}_i||\) represents the Euclidean distance between data point \(\mathbf{x}\) and centroid \(\boldsymbol{\mu}_i\). K-means offers simplicity and efficiency, making it a valuable tool for data analysis, image processing, and more. We utilize this clustering technique to efficiently discretize the continuous state space $\mathcal{S}$ into $K$ number of clusters for additional cost penalization by measuring their sparsity within the cluster.

\section{Simple Sparsity-based Cost Penalty} \label{sec:methodology}
As outlined in Section \ref{sec:preliminaries}, DICE algorithms represent a model-free policy gradient method that estimates marginalized visitation ratios between the distributions of data and the learning policy. Because this requires an estimation with a finite set of data, estimation error is inevitable. This observation highlights a significant trend: while numerous methods concentrate on mitigating extrapolation using \emph{f-divergence} metrics between the distributions of the learning policy and the data, there's a noticeable absence of studies addressing interpolation errors stemming from discontinuities in the inner distributions of the data.

Hence, relying on previous naive OPE methods for learning can result in safety breaches primarily due to interpolation errors, as illustrated in \cite{lee2022coptidice}. Given the considerable challenges of ensuring uncertainty in offline and nonlinear settings, a straightforward yet effective approach is to introduce additional conservatism. Specifically, one may alleviate safety violations in naive OPE via extra conservatism for estimation errors. However, one must take a holistic and sophisticated approach to measure such conservatism by avoiding over-conservatism but effectively addressing estimation errors. Thus, our motivation for applying a stronger penalty in sparser regions is twofold: 1) estimation errors by interpolation are more prone to occur in sparse data regions, and 2) the need to avoid over-conservatism to preserve the potential for the highest return. In essence, our aim is to introduce a relative degree of conservatism by categorizing the given data distributions into regions of confidence and regions of lesser confidence. This effectiveness is demonstrated in Section \ref{sec:experiments} through a comparison with naive penalty methods (i.e., simply multiplying an integer for penalization).

\subsection{Tabular settings}
\begin{figure}[!h]
\centering
\begin{tabular}{c c}
    \includegraphics[width=0.23\textwidth]{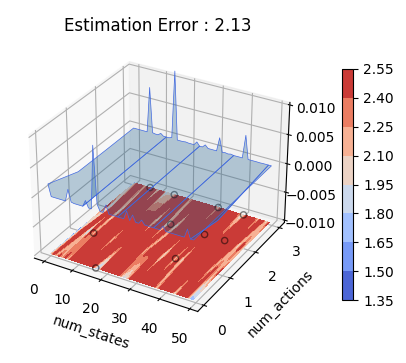}    &
    \includegraphics[width=0.23\textwidth]{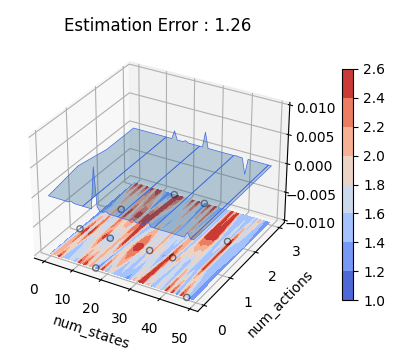}    \\
    (a) $\mathcal{N}=10$ & (b)  $\mathcal{N}=100$ \\
\end{tabular}
\caption{The estimation error, $C_{\text{true}}/C_{\text{est}}$, was assessed for two different data volumes, and the discrepancy, $C_{\text{true}} - C_{\text{est}}$, is depicted as a blue surface. The color gradient below the plot indicates the level of penalty for the corresponding state-action pairs, with specific values for conservatism shown in the adjacent bar. Panel (a) shows the estimation error with a limited number of trajectories $\mathcal{N}$, and panel (b) demonstrates the scenario with a moderate number of trajectories, both yielding errors.}
\label{fig_1}
\end{figure}

\begin{figure*}[t]
    \centering
    \begin{tabular}{c c}
        \multicolumn{2}{c}{\includegraphics[width=0.5\textwidth]{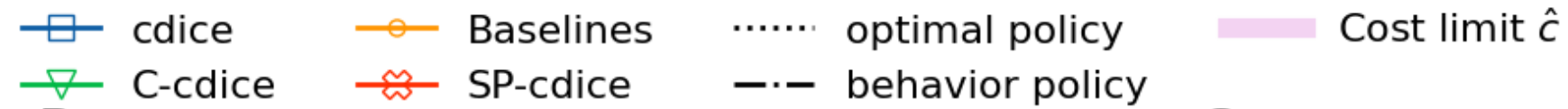}}\\
        \includegraphics[width=0.475\textwidth]{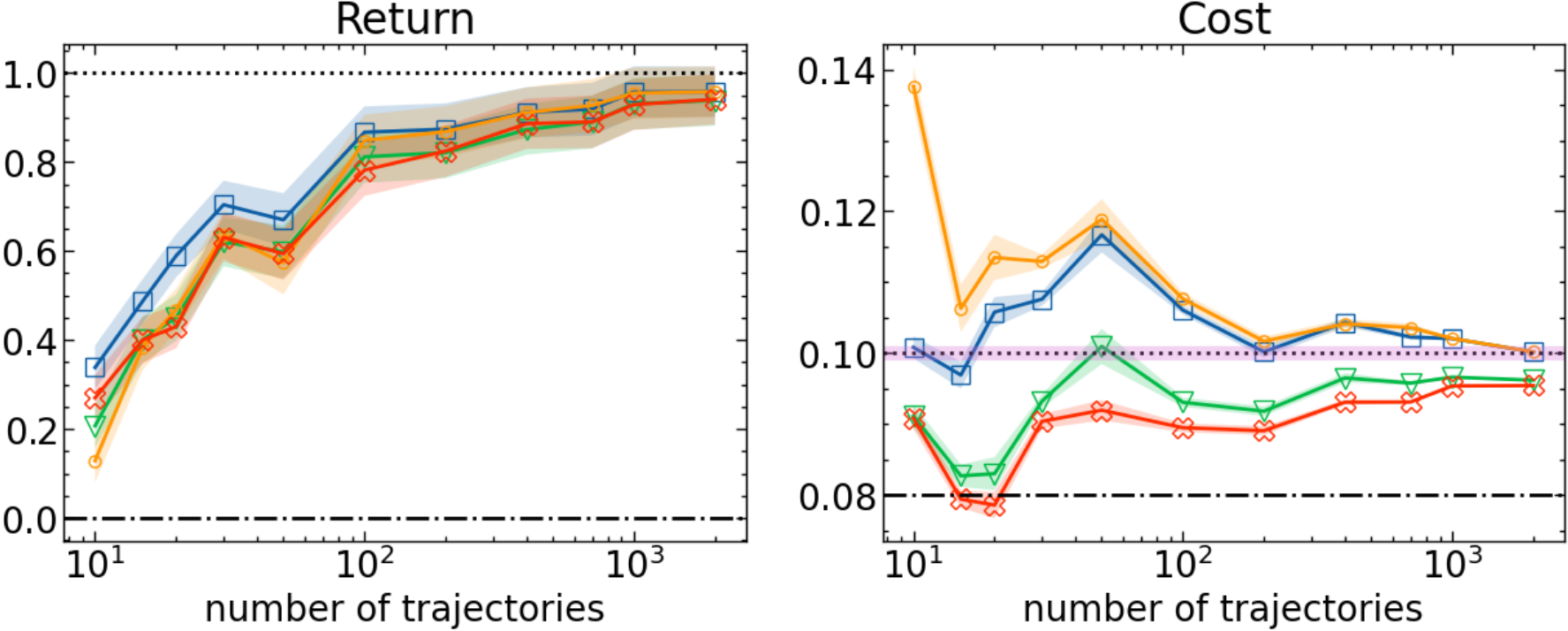}    &
        \includegraphics[width=0.475\textwidth]{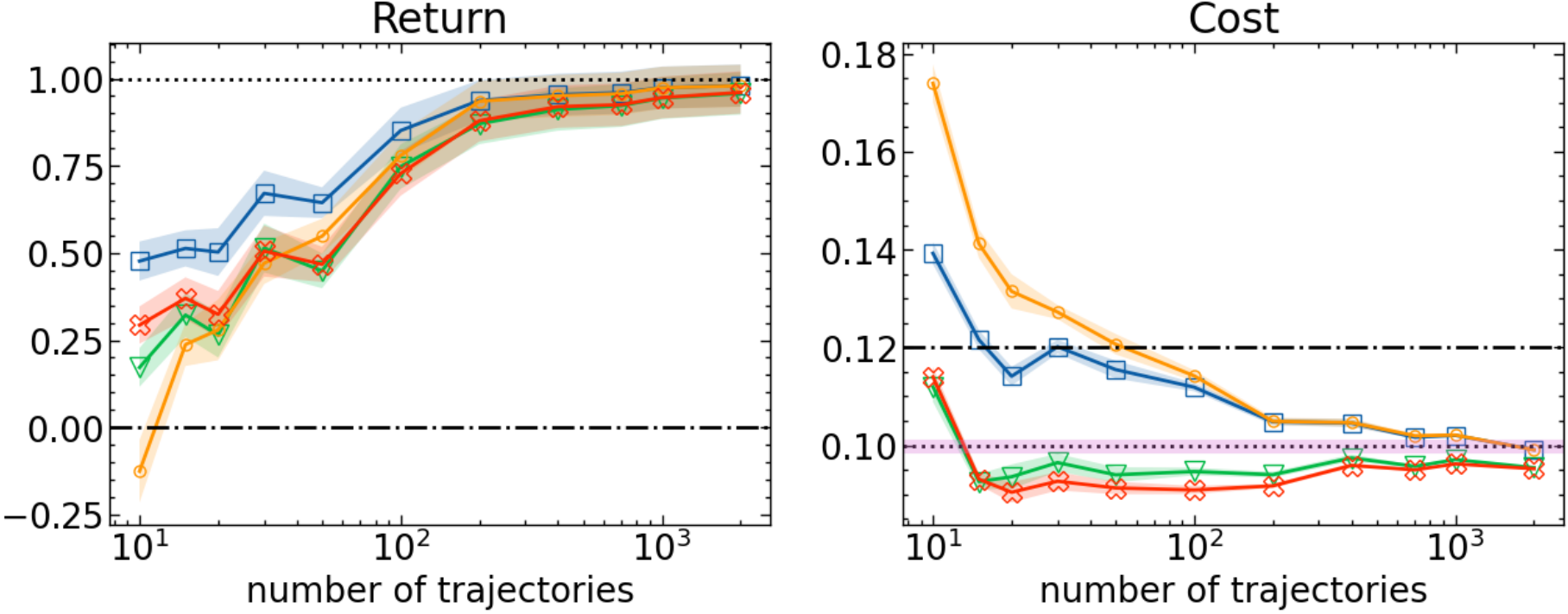}    \\
        (a) Cost Satisfying Dataset & (b) Cost-violating Dataset \\
    \end{tabular}
    \caption{A random CMDP with a cost limit of $c=0.1$ was tested using 10 seeds, varying the number of trajectories to compose datasets for better environmental understanding. Two datasets were created: one meeting the cost limit and the other not. The solid line represents the average, and the shaded area shows the standard deviation. Overall, SP-cdice and conservative COptiDICE consistently satisfied the constraint even with fewer trajectories, outperforming other methods in ensuring return and safety. This highlights the effectiveness and efficiency of our novel safety conservatism approach compared to traditional methods.}
    \label{fig:tabular_exp}
\end{figure*}

Through the empirical studies, we found that the number of visits was effective in tabular settings. Emerged as a measure of the \emph{'exploration'}, it was initially incorporated into Reinforcement Learning as a term \emph{bonus} in \cite{rezaeifar2022offline}. \emph{Bonus}, $b(s,a)$, is a function of state-action, and it is defined as $b(s,a) \propto 1/\sqrt{n(s,a)}$, where $n(s,a)$ representing the accumulated visits to state-action pairs. By incorporating this measure, the policy update involves subtracting a 'bonus' from the expected return: $\pi_{k+1} = \argmax_{\pi} \langle Q_{\pi} - b(s,a) \rangle$. This reduction in expected values for less-explored (uncertain) state-action pairs helps restrict out-of-distribution actions. In constrained offline RL, we demonstrate that this 'bonus' is also valuable for identifying potentially more hazardous state-action pairs in tabular settings. In essence, we repurpose the bonus for cost overestimation, introducing a hyperparameter $\alpha > 0$. To connect the notations to that of COptiDICE, we use $\omega^*_{tabular}(s,a)$ instead of \emph{bonus}, and formulation becomes:

\begin{equation}
    \mathbb{E}_{(s,a) \sim d^D} [C_k(s,a)\omega(s,a)\omega^*_{tabular}(s,a)] \leq \hat{c}_k.
\end{equation}
where $\omega^*_{tabular}(s,a) := [\alpha/\sqrt{n(s,a)} + 1] \geq 1$. 
Prior to imposing this constraint on \Cref{eqn:coptidice}, we illustrate the estimation error of naive COptiDICE (without $\omega^*_{\text{tabular}}(s,a)$) in \Cref{fig_1} which is Random MDP with composition of ($\mathcal{S} \times \mathcal{A} \subset \mathbb{R}^{50 \times 4}$). Each small dot, denoting the top 10 estimation errors, corresponds to relatively high conservatism (indicated by red-tone colors, with specific values shown in the adjacent bar). This implies that less-explored regions are more susceptible to errors. For experiment results in a tabular setting, please see \Cref{fig:tabular_exp}.

\subsection{Continuous State Space}
While the bonus, $b(s,a)$, proved valuable in simple discrete tabular settings, its applicability wanes in the case of continuous state-action spaces. Counting state-action visitation, $n(s,a)$, in continuous contexts becomes pointless due to the infinite number of possible choices, rendering it an ineffectual measure. 
Additionally, the challenge of scalability arises when discretizing state-action spaces to continue utilizing bonus metrics. This involves the need to partition spaces effectively, which can become infeasible for tasks with high-dimensional state and action spaces. To overcome this challenge, \cite{rezaeifar2022offline} advocated the use of counting by density models \cite{bellemare2016unifying}, \cite{pmlr-v70-ostrovski17a} or counting after hash function \cite{tang2017exploration} to gauge the extent of the exploration. Nonetheless, both of these methods entail the development of an additional model, introducing an added layer of complexity. However, our approach offers a novel solution as a simple pre-processing stage before training begins. In particular, by measuring sparsity exclusively from the provided static dataset, it circumvents the necessity for an extra model or optimization problem, as mandated by prior methodologies.

\begin{figure*}[t]
    \centering
    \begin{tabular}{c c c c c}
        \multicolumn{4}{c}{CartPole Data Sparsity Visualization}\\
        \hline\hline
        \includegraphics[width=0.225\linewidth]{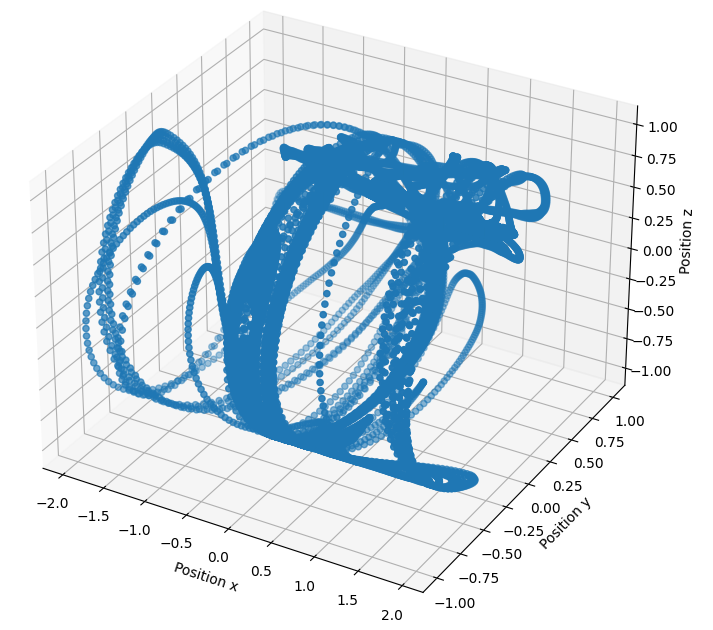}    &
        \includegraphics[width=0.225\linewidth]{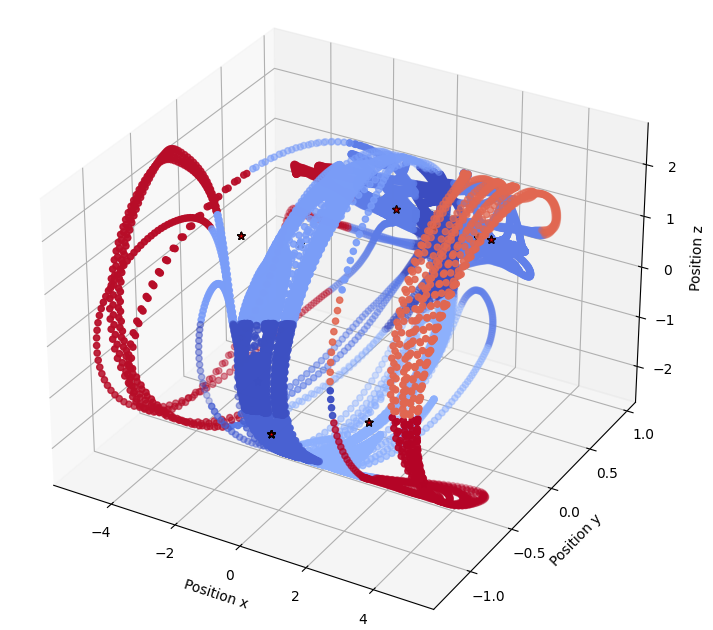}   &
        \includegraphics[width=0.225\linewidth]{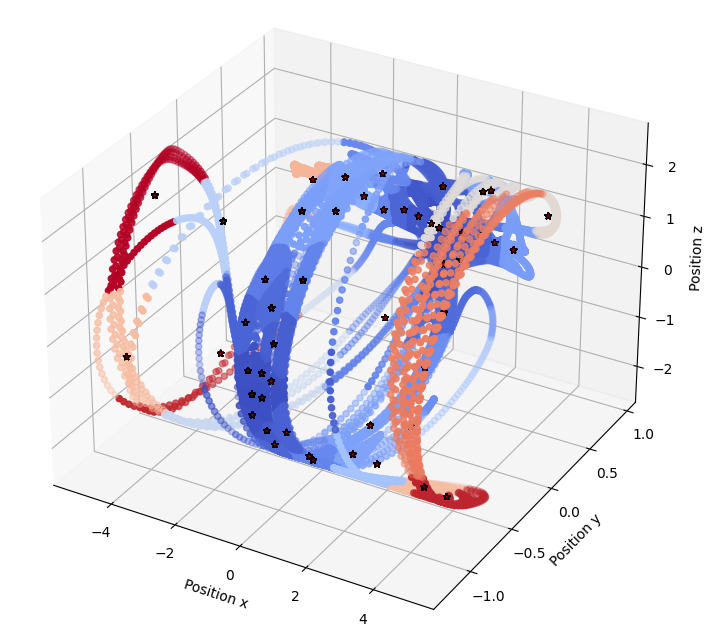}   &
        \includegraphics[width=0.225\linewidth]{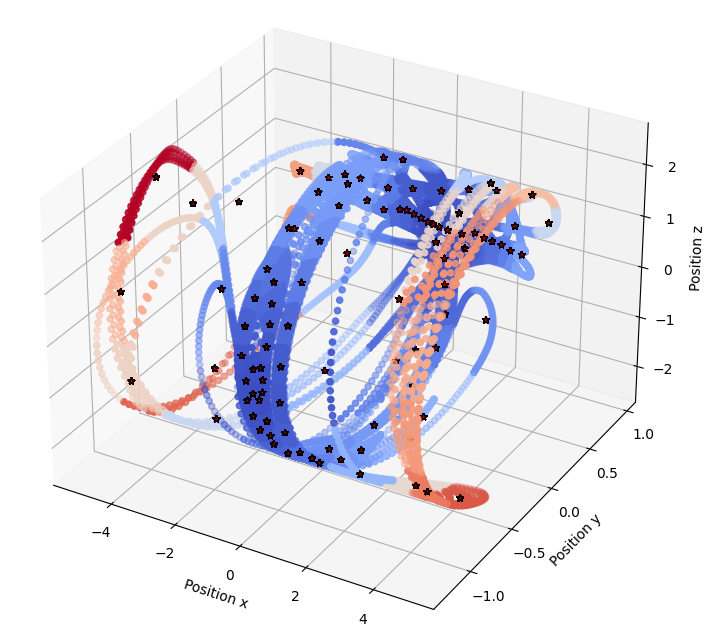}  \\
        
        (a) Original Geometry  & (b) $K = 10$ & (c) $K = 50$ & (d) $K = 100$    \\
    \end{tabular}
    
    \caption{This figure illustrates the positional state spaces $\mathcal{S} \subset \mathbb{R}^{3}$ of the Cartpole environment. (a) shows the state data distribution, while part (b) displays penalties for state clusters with 10 clusters. Reddish tones indicate higher anti-confidence levels, and little dots are centroids of each cluster. Parts (c) and (d) provide more detailed analyses with $K$ = 50 and 100, respectively.
    }
    \label{fig:sparsity_visualization}
\end{figure*}

\subsubsection{Sparsity Measure}
As part of the pre-training process, we can measure point-wise nearest numerical deviations to quantify sparsity. While this approach intuitively gauges how closely a data point is positioned to its neighbors, it demands significant computational resources to handle high-dimensional datasets with large volumes. This computational challenge renders the use of such a sparsity metric less appealing compared to previous methods, primarily due to its impracticality.

 However, in this paper, we address this issue by grouping data into $\mathcal{K}$ number of clusters using $K$-means clustering, denoted as $\mathcal{S} \subset \mathcal{D}^{n \times m} \Longrightarrow \mathcal{S}_k \subset \mathcal{D}_k^{n_k \times m}$, where $\mathcal{D}$ is the given dataset. By discretizing the continuous state space into $K$ clusters, we relax the computational burden and measure numerical deviations of surrounding data points with respect to their respective centroids. In particular, we refrain from dividing the data dimensions into uniform sectors, opting instead to discretize them into clusters. This allows us to consider the respective sparsity of data instead of discretizing the whole data regime into uniform sectors.
 
 This method stands as an alternative penalty measure for conservative interpolation. We notably employ z-score standardization to ensure that the sparsity penalty, denoted as $\omega^*(s)$, operates within similar numerical ranges across the dataset $\mathcal{D}$ and different domains while preserving its geometric properties. \Cref{fig:sparsity_visualization} depicts the intact geometry of data shape and the corresponding penalties assigned to clusters based on their sparsity (where the red color indicates a higher penalty). In contrast to the tabular method, $\omega^*(s)$ is solely a function of the state, as state-action counts do not provide significant benefits, as argued by \cite{tang2017exploration}.

The following formula is used to calculate sparsity for each of the $k^{th}$ clusters:
\begin{equation} \label{eqn:normalized_penalty_score}
    \omega_k(s) = \text{Normalize} \biggl[ \frac{\sum_i^{\mathcal{N}}(\mathcal{C}_k -  d_k^i)^2}{\mathcal{N}m}\biggr]
\end{equation}
In this formula, $\mathcal{N}$ represents the number of data points in each cluster, $\mathcal{C}_k$ is the centroid of the cluster, $d_k^i$ denotes the $i$-th element within the $k^{th}$ cluster, and $m$ is the state dimension. This calculation provides a normalized measure of sparsity for each given data. 
Building on the methodology outlined in \cite{lee2022coptidice}, we utilized the following conservatism metric for each batch update:

\begin{equation} \label{eqn:final_sparsity_penalty}
    \omega^*(s) := \text{softmax}\biggl[ \omega_k(s) \biggr] * \text{batch\_size}, \quad \omega^*(s) \geq 1
\end{equation}

The above definition allows for an exponential penalty, applied to sparse regions using the normalized conservatism score in \Cref{eqn:normalized_penalty_score} for the data in the k-th cluster. In contrast, approaches like \cite{chow2015risk} and \cite{lee2022coptidice} have tackled this issue by employing an additional bi-level cost upper bound optimization through perturbation. Meanwhile, \cite{bellemare2016unifying}, \cite{pmlr-v70-ostrovski17a}, and \cite{tang2017exploration} require the use of an additional model for cost overestimation, which are also susceptible to weaknesses in hyperparameter tuning. In the following section, we evaluate the effectiveness of the sparsity-based penalty through empirical studies. We provide our pseudo-code in \Cref{alg:sp_dice}.

\begin{algorithm}[t]
    \caption{Sparsity-based Safe Reinforcement Learning}\label{alg:sp_dice}
    \begin{algorithmic}[1]
        \REQUIRE Dataset $\mathcal{D}$, Constrained RL algorithm Alg$(\cdot)$
        \REQUIRE number of clusters $k$, minibatch $n$
                \STATE \texttt{/* Sparsity Measurement */}
                \STATE Compute the clusters with centroids using $k$-Means clustering as in \Cref{eqn:clustering}
                \STATE Compute the sparsity, $\omega^*(s)$, of each cluster as in \Cref{eqn:final_sparsity_penalty}
            \STATE \texttt{/* RL Training Loop */}
          \WHILE{not done}
              \FOR{each minibatch $\mathcal{D}_n$ in $\mathcal{D}$}
                    \STATE $C^*(s,a) = C(s,a) \cdot \omega^*(s)$ where $(s,a) \in \mathcal{D}_n$
                    \STATE Train Alg$(\mathcal{D}^*_n)$ where $\mathcal{D}^*_n = (S, A, S', C^*, R)$
                \ENDFOR
        \ENDWHILE
    \end{algorithmic}
\end{algorithm}

\section{Experiments} \label{sec:experiments}
In this experiment, we performed a comparative analysis between our sparsity-based penalization approach and the perturbation method employed by \cite{lee2022coptidice}, using their respective algorithms. Given the limited existing research in the field of constrained offline Reinforcement Learning, we selected Constraints Penalized Q-learning (CPQ) \cite{DBLP:journals/corr/abs-2107-09003} and Constrained Offline Policy Optimization via Stationary DIstribution Correction Estimation (COptiDICE) \cite{lee2022coptidice} as our comparative methods. For our experimental setup, we opted for the Random CMDP environment as utilized by \cite{lee2022coptidice} and \cite{laroche2019safe} for discrete state and action spaces. Additionally, we selected four cost-violating environments from benchmarks provided by Datasets for Safe Reinforcement Learning (DSRL) \cite{liu2023datasets}, using RWRL suite \cite{dulacarnold2020realworldrlempirical} for continuous state-action spaces. 

\subsection{Discrete: Random CMDP}
Following \cite{lee2022coptidice} and \cite{laroche2019safe}, we created a random Constrained Markov Decision Process (CMDP) with the following parameters for our experiments: 
\begin{itemize}
    \item \textbf{Random CMDP $(\mathcal{S} \subset \mathbb{R}^{50}, \mathcal{A} \subset \mathbb{R}^4)$:} A goal was placed in rarely-visited states by assigning a reward of 1, while all other states received a reward of 0. The cost function, $C(s, a)$, was generated randomly with a cost threshold $\hat{c} = 0.1$.
\end{itemize}
The transition probabilities were randomly generated with a connectivity of 4, and $|A| = 4$. 

\subsubsection{Evaluation}
With both optimal and behavior policies, we compare our SP-cdice (SParsity-based conservative COptiDICE) algorithm, described in \Cref{alg:sp_dice}, with its naive and conservative versions. For comprehensive evaluation, we also include the policy iteration algorithm for this discrete environment setting. Running 10 seeds, our SP-cdice demonstrates performance close to the conservative COptiDICE while remaining less computationally expensive. This suggests that SP-cdice can be applied to other settings, such as continuous environments, as discussed below.

\subsection{Continuous: CartPole}
Here, we showcase the performance of our approach in a continuous setting. Specifically, we test it in the Cartpole environment from RealWorldRL (RWRL) \cite{dulacarnold2020realworldrlempirical}, as in \cite{lee2022coptidice}:

\begin{itemize}
    \item \textbf{Cartpole $(\mathcal{S} \subset \mathbb{R}^{4}, \mathcal{A} \subset \mathbb{R}^1)$:} The cart aims to keep the pole upright without falling, while being constrained to stay within specified regions. The action is the force applied to the body of the cart.
\end{itemize}

Our primary focus is on validating the efficacy of sparsity-based conservatism. To achieve this, we apply a multiplier to the corresponding costs, penalizing the cost in sparse regions at a higher scale compared to dense regions. This approach suggests a relative degree of conservatism between sparse and dense data regions to effectively address interpolation errors. In contrast, the naive penalty approach involves multiplying a constant across all costs in the data for overestimation.

\subsubsection{Evaluation}

\begin{figure}[t]
    \centering
    \includegraphics[width=0.45\textwidth]{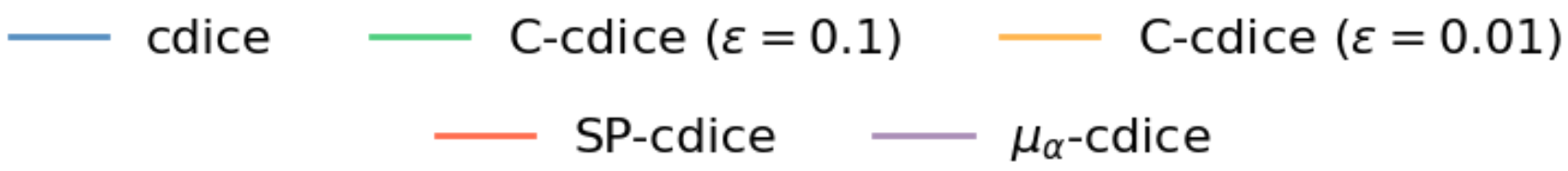}\\
    \includegraphics[width=0.45\textwidth]{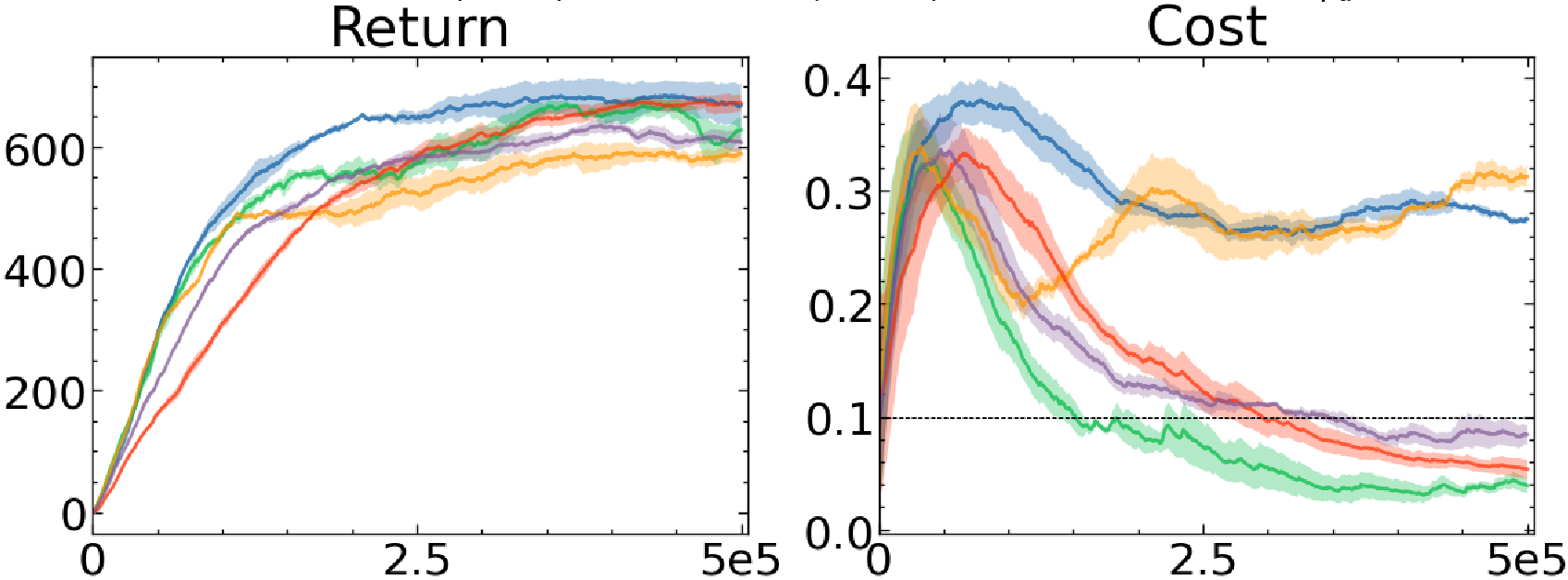} \\
    Cartpole ($\alpha=10$)
    \caption{We experimented with our algorithm in the continuous domain (Cartpole). Parameter $\alpha$ was chosen for the naive penalty method to ensure compliance with the cost limit. Our approach outperformed or same as non-conservative methods in terms of performance while adhering to the cost limit. In contrast, the naively constrained algorithm achieved lower returns, underscoring the effectiveness of our novel relative conservatism approach using sparsity measures. This demonstrates the applicability and superiority of our method in continuous settings compared to existing approaches.}
    \label{fig:enter-label}
\end{figure}

The value of $\alpha$ was selected to ensure the curve stays below the cost limit through trial and error, demonstrating our approach's relative degree of conservatism. This is validated by our return results, showing that our method achieves higher returns while remaining under the cost limit. While this highlights the effectiveness of our approach in capturing a relative degree of conservatism, similar to the tabular setting, our approach shows that it can achieve better returns than the conservative COptiDICE, which requires parameter tuning ($\epsilon$).

\section{Related Works}
Constrained reinforcement learning (RL) has been an active research area, with significant contributions from various studies. \cite{achiam2017constrained} introduced the trust-region approach for constrained RL, ensuring monotonic improvement with safety guarantees. Building on this, \cite{cho2024constrained} extended its theoretical guarantees to the constrained meta-learning setting. \cite{chow2015risk} employed Conditional Value-at-Risk (CVaR) as a constraining metric, while \cite{alshiekh2018safe} implemented a shielding mechanism. \cite{ghavamzadeh2016safe} conducted a theoretical study on regret in constrained settings, and \cite{turchetta2016safe} utilized Gaussian Processes (GP) for robust probabilistic uncertainty analysis.

In the context of offline RL, several notable works have emerged. \cite{yu2020mopo} presented a model-based approach without constraints, incorporating an uncertainty penalty to remain within the data regime. \cite{cho2024out} enabled robust out-of-distribution exploration by leveraging causal knowledge and novel network architectures to detect erroneous predictions. \cite{xu2022constraints} proposed a value-based approach that also constrains out-of-distribution actions. Finally, model-free approaches such as \cite{lee2021optidice, lee2022coptidice} estimated stationary distributional correction to bridge offline gradient estimators to online settings.

\section{Conclusion}
We presented a novel approach to sparsity-based safe reinforcement learning, leveraging the power of clustering to introduce an additional layer of conservatism in offline RL settings. By testing our SP-cdice (SParsity-based conservative COptiDICE) algorithm in both discrete and continuous environments, we have demonstrated its effectiveness and computational efficiency. Our SP-cdice algorithm performs comparably to conservative COptiDICE, achieving substantial returns while being highly generalizable as a preprocessing step for data, indicating its potential for broader applications.

\bibliography{refs}
\bibliographystyle{IEEEtran}

\end{document}